\documentclass[conference]{IEEEtran}
\IEEEoverridecommandlockouts
\usepackage{cite}
\usepackage{amsmath,amssymb,amsfonts}
\usepackage{algorithmic}
\usepackage{graphicx}
\usepackage{textcomp}
\usepackage{xcolor}
\usepackage{multicol}
\usepackage[a4paper, margin=1in]{geometry}
\usepackage{amsmath}
\usepackage{booktabs}
\usepackage{graphicx} 
\usepackage{float}
\usepackage[ruled,vlined]{algorithm2e}
\usepackage{placeins}
\usepackage{booktabs}
\usepackage{subfigure}
\usepackage{subcaption}
\newcommand\blfootnote[1]{%
  \begingroup
  \renewcommand\thefootnote{}\footnote{#1}%
  \addtocounter{footnote}{-1}%
  \endgroup
}

\graphicspath{ {./Images/} }

\def\BibTeX{{\rm B\kern-.05em{\sc i\kern-.025em b}\kern-.08em
    T\kern-.1667em\lower.7ex\hbox{E}\kern-.125emX}}
    
\begin{document}

\title{Path Planning in a dynamic environment using Spherical Particle Swarm Optimization
}
\author{Mohssen E. Elshaar$^{1,*}$, Mohammed R. Elbalshy$^{1,*}$, A. Hussien $^{2,\dagger}$ and Mohammed Abido$^{2,\dagger,\ddagger}$\\
\footnotesize{$^1$Department of Aerospace Engineering, $^2$Department of Electrical Engineering}\\
\footnotesize{$^\dagger$Interdisciplinary Research Center for Sustainable Energy Systems (IRC-SES)}\\
\footnotesize{$^\ddagger$SDAIA-KFUPM Joint Research Center for Artificial Intelligence}\\
\footnotesize{King Fahd University of Petroleum and Minerals (KFUPM), Dharan, Saudi Arabia}
}

% \thanks{\begin{flushleft}
%     \indent Emails: {\texttt{\{g202309590, g202216780, mabido\}@kfupm.edu.sa}} according to authors' order.
% \end{flushleft}}
% }

\maketitle
\blfootnote{$^*$Equal Contribution.\\
Emails:{\{g202309590,g202216780,husseina,mabido\}@kfupm.edu.sa}
}

\begin{abstract}
    Efficiently planning an Unmanned Aerial Vehicle (UAV) path is crucial, especially in dynamic settings where potential threats are prevalent. A Dynamic Path Planner (DPP) for UAV using the Spherical Vector-based Particle Swarm Optimisation (SPSO) technique is proposed in this study. The UAV is supposed to go from a starting point to an end point through an optimal path according to some flight criteria. Path length, Safety, Attitude and Path Smoothness are all taken into account upon deciding how an optimal path should be. The path is constructed as a set of way-points that stands as re-planning checkpoints. At each path way-point, threats are allowed some constrained random motion, where their exact positions are updated and fed to the SPSO-solver. Four test scenarios are carried out using real digital elevation models. Each test gives different priorities to path length and safety, in order to show how well the SPSO-DPP is cabable of generating a safe yet efficient path segments. Finally, a comparison is made to reveal the persistent overall superior performance of the SPSO, in a dynamic environment, over both the Particle Swarm Optimisation (PSO) and the Genetic Algorithm (GA). The methods are compared directly, by averaging costs over multiple runs, and by considering different challenging levels of obstacle motion. SPSO outperformed both PSO and GA, showcasing cost reductions ranging from 330\% to 675\% compared to both algorithms.
\end{abstract}


\section{Introduction}
Unmanned aerial vehicles (UAVs), commonly referred to as drones, are progressively gaining prevalence and proving their utility across diverse domains and industries, ranging from agriculture and surveillance to disaster response and aerial photography. One important factor that is crucial to the efficient use of UAVs during deployment is path planning which involves the systematic determination of the optimal route a drone or a UAV should follow to achieve its mission objectives while adhering to various constraints and minimizing risks \cite{phung2017enhanced, phung2019system}. Key considerations in UAV path planning include obstacle avoidance, energy efficiency, real-time decision-making, and adaptation to changing conditions. Researchers and engineers have developed a plethora of approaches to tackle UAV path planning challenges, from traditional methods like $A^*$ search and Dijkstra's algorithm to advanced techniques like Rapidly-exploring Random Trees ({RRT}) and Genetic Algorithms ({GA}).\\

The criteria for optimal path planning extend to diverse objectives depending on the application, involving the maximization of detection probability, flight time minimization or Pareto solutions achievement for the navigation of multi-objective problems \cite{phung2020motion, lin2009uav, yin2018offline}. Furthermore, the intended trajectory must adhere to relevant safety and feasibility limits within the operational environment of the UAV. Safety considerations include ensuring that the path can safely steer the UAV past environmental risks such as barriers. The path must be feasible in terms of altitude, flight time, fuel consumption, ascending angle and turning rate. As a result, path planning aimed at increasing safety by permitting collision-free and practicable UAV motion remains a constantly difficult task. Various path planning strategies have been proposed including approaches such as graph search, cell decomposition, potential field, and nature-inspired algorithms \cite{beard2002coordinated, mclain2005coordination, eppstein1998finding, pettersson2006probabilistic, lin2017sampling, barraquand1992numerical, chen2016uav, di2015potential}.

Lately, nature-inspired approaches are increasingly favored in UAV path planning due to their efficiency in accommodating dynamic UAV constraints and seeking global optima in intricate scenarios. Various algorithms, drawing inspiration from nature, are tailored for UAV path planning, including cuckoo search \cite{song2020parallel}, genetic algorithm (GA) \cite{roberge2012comparison, roberge2018fast}, differential evolution (DE) \cite{fu2013route}, artificial bee colony (ABC) \cite{xu2010chaotic}, ant colony optimization (ACO) \cite{yu2018aco}, and particle swarm optimization (PSO) \cite{phung2017enhanced, roberge2012comparison, yu2018aco,fu2011phase}. PSO, in particular, finds widespread use and has seen several variants developed. PSO is a population-based algorithm inspired by bird flocking and fish schooling behavior that embodies swarm intelligence through cognitive and social coherence \cite{kennedy2006swarm}. These qualities allow each particle to navigate the solution space based on individual and collective experiences, which differs from standard evolutionary operators such as mutation and crossover. As a result, PSO efficiently identifies global solutions with stable convergence, outperforming alternative nature-inspired algorithms \cite{gaing2003particle}. PSO variants proposed in \cite{das2020multi, zhang2013robot, wei2008theta, hoang2018angle, zhang2014adaptive} share a population-based structure but differ in search space representation and particle-encoded solutions.
 
Real-world problems are often dynamic, involving changing global and local optimal solutions that vary over time. Dynamic Path Planners were designed based on improved Simulated Annealing (SA) \cite{shi2023dynamic}, Neural Networks (NN) \cite{decastro2023dynamic}, Reinforcement Learning \cite{mpprl-uav} and Fuzzy Logic \cite{hentout2023review}.
Some studies argue that Evolutionary Algorithms (EAs) are well-equipped to handle these challenges, evidenced by their success in dynamic problems \cite{branke2001evolutionary, branke2003designing, branke2012evolutionary}. From these algorithms, PSO has been employed in a variety of path planning issues in dynamical contexts, including \cite{blackwell2007particle, alaliyat2019path, parrott2004particle}. 
A new algorithm, known as Spherical Vector-based Particle Swarm Optimization ({SPSO}), is introduced in \cite{spso2021} to tackle the problem of path planning for UAVs in complex and static threat-filled environments. It begins by creating a cost function that turns the path planning problem into an optimization task, considering UAV operational requirements and safety constraints. SPSO is then applied to efficiently search the UAV's configuration space to find the optimal path. This is done by linking particle positions to critical flight parameters, including speed, turning angle, and climb/dive angle. SPSO was tested exclusively in a static environment, and its performance in a dynamic environment remains unexplored. In essence, this work focuses on:

\begin{enumerate}
    \item Introducing Spherical Particle Swarm Optimization Dynamic Path Planner (SPSO-DPP).
    \item Investigating the efficiency of SPSO-solver within a dynamic setting characterized by the presence of randomly moving obstacles.
    \item Comparing the SPSO-DPP performance with PSO and GA performances in a dynamic environment.
\end{enumerate}

\section{Methodology}
\subsection{Problem formulation}
An optimisation problem is used to formulate the UAV path planning problem. The constraints are obtained from the UAV flying constraints, and the path objective function is defined as the path length \cite{spso2021}.
\subsubsection{Path Cost}
For many aerial picture and mapping applications, the primary optimum criterion is to minimise the path length. The set of $n$ way-points that the UAV must travel through represents the flight path. Each co-ordinate $(x_{i}, y_{i}, z_{i})$ describes a way-point $\vec{P}_{i}$ in detail. The separation between every two way-points is:
\[ d_{i} = \|\vec{P}_{i+1} - \vec{P}_{i}\|\]
The cost $F_1$ that needs to be minimized is the summation of all distances $d_{i}$, where $i \in \{1,2,3, .. , n-1\}$.
\begin{equation}\label{patheqn}
    F_{1} = \sum_{i=1}^{n-1} d_{i}
\end{equation}
\subsubsection{Safety Constrain}
In order to protect the UAV from any threats resulting from impediments in its operating region, the planned route needs to be carefully developed. This means that a safety constraint must be included in the planning phase. A set that represents all possible threats, $K$, is defined. It is assumed that every danger is confined within a cylindrical container, and that each threat's projection has a radius of $R_k$ and a central coordinate $C_k$.
The cost associated with potential threats for a given path segment $\Delta\vec{P}_{i} = \vec{P}_{i+1} - \vec{P}_{i}$ is proportional to the distance, $d_k$, between the segment and the central coordinate $C_k$ of each danger. The threat cost $F_2$ is calculated along the way-points $P_{i}$ with regard to the obstacle set $K$ in the following method to account for the physical dimensions of the UAV, which are defined by its diameter $D$ and the critical safety threshold $S$ that defines the danger zone for probable collisions:

\begin{equation}\label{safetyeqn}
    F_2 = \sum_{i=1}^{n-1} \sum_{k=1}^{K} T_{k} \Delta\vec{P}_{i},
\end{equation}
where
{
\scriptsize
\begin{equation}
    T_k =
    \begin{cases}
        0 & \text{if } d_k > S + D + R_{k} \\
        (S + D + R_{k}) - d_{k} & \text{if } D + R_{k} < d_{k} \leq S + D + R_{k} \\
        \infty & \text{if} d_{k} \leq D + R_{k}.
    \end{cases}
\end{equation}
}
\subsubsection{Altitude Constrain}
The flying altitude is frequently restricted during operation between the minimum and maximum heights, the two designated extrema. For instance, the camera must capture visual data at a particular resolution and field of view for surveying and search applications, which limits the flying altitude. Let $h_{\text{min}}$ and $h_{\text{max}}$ stand for the lowest and highest heights, respectively. For each way-point $P_{i}$, the altitude cost is calculated as:
\begin{equation}
H_{i} =
\begin{cases}
    \left| h_{i} - \frac{h_{\text{max}} + h_{\text{min}}}{2} \right|, & \text{if } h_{\text{min}} \leq h_{i} \leq h_{\text{max}} \\
    \infty, & \text{otherwise}
\end{cases}
\end{equation}
In this representation, the flying height above the ground is shown by $h_{i}$. It is made so that out-of-range values are penalised and the average height is maintained by $H_{i}$. The altitude cost is obtained by adding together all of the way points' $H_{i}$:
\begin{equation}\label{altiteqn}
    F_3 = \sum_{i=1}^{n} H_{i}
\end{equation}

where $n$ is the total number of way-points.
\subsubsection{Smoothness Constrain}
The smoothness cost function assesses the rates of turning and climb, both of which are vital for creating workable flight trajectories.
The turning angle, represented as $\phi_{i}$, signifies the angle between two successive segments of the path. This angle is determined by projecting the vectors originating from point $P'_{i}$ to $P'_{i+1}$ and from point $P'_{i+1}$ to $P'_{i+2}$ onto the horizontal plane denoted as $O_{xy}$.
\begin{equation}
    \phi_{ij} = \arctan\left(\frac{\| \overrightarrow{P'_{i}P'_{i+1}} \times \overrightarrow{P'_{i+1}P'_{i+2}} \|}{\overrightarrow{P'_{i}P'_{i+1}} \cdot \overrightarrow{P'_{i+1}P'_{i+2}}}\right)
\end{equation}
The angle between the path section from $P_i$ to $P_i+1$ and its projection into the horizontal plane - that is, the line from $P'_i$ to $P'_i+1$ — is known as the angle of climb, indicated by the symbol $psi_i$. The following formula can be used to compute this angle:

\begin{equation}
    \psi_{ij} = \arctan\left(\frac{z_{i+1} - z_{i}}{\|\overrightarrow{P'_{i}P'_{i+1}}\|}\right)
\end{equation}

The calculation of the smoothness cost is as follows:
\begin{equation}\label{smootheqn}
    F_4 = a_1 \sum_{i=1}^{n-2} \phi_{i} + a_2 \sum_{i=1}^{n-1} \left( \psi_{i} - \psi_{i-1}\right)
\end{equation}

Here, $a_1$ and $a_2$ represent the penalty coefficients associated with the turning and climbing angles, respectively.

\subsubsection{Accumulative Cost Function}

Taking into account considerations related to optimizing the path, ensuring safety, and meeting altitude and smoothness constraints, we can formulate the overall cost function as follows:
\begin{equation}\label{finalcost}
F(X_j) = \sum_{k=1}^{4} b_k F_k(X_j)
\end{equation}
Here, each $b_k$ serves as a weight coefficient, and $F_1$ through $F_4$ represent the costs associated with different aspects: path length (\ref{patheqn}), safety cost (\ref{safetyeqn}), flight altitude (\ref{altiteqn}), and smoothness (\ref{smootheqn}). The decision variable, denoted as $X_j$, encompasses a list of $n$ way-points from the starting point to the end point, $P_{ij} = (x_{ij}, y_{ij}, z_{ij})$, satisfying the condition $P_{ij} \in O$, where $O$ signifies the operational space for the UAV. With these definitions, the cost function $F$ becomes fully defined, and the optimization problem (\ref{finalcost}) is aimed to be solved for the optimum $X^{\ast}$ in which the $F(X^{\ast})$ is minimum.

In the SPSO approach, each flight path is translated into a collection of vectors, each specifying how the UAV moves from one waypoint to another. These vectors are described in a spherical coordinate system, consisting of three key elements: the magnitude $\rho \in [0, \text{path length}]$, the elevation angle $\psi \in [-\pi/2, \pi/2]$, and the azimuth angle $\phi \in [-\pi, \pi]$. A flight path indexed as $j$ and comprising of $N$ nodes can be represented as a hyper-spherical vector with $3N$ dimensions:
\begin{equation}
    \begin{aligned}
        \Omega_{j} &= (\rho_{1j}, \theta_{1j}, \phi_{1j}, r_{2j}, \theta_{2j}, \phi_{2j}, \\ &\dots, r_{Nj}, \theta_{Nj}, \phi_{Nj});  \text{where } N = n - 2  
    \end{aligned} 
\end{equation}

When we characterize a particle's position as $\Omega_{i}$, the particle's velocity is defined by an incremental vector:
\begin{equation}
    \begin{aligned}
        \Delta\Omega_{j} &= (\Delta r_{1j}, \Delta \theta_{1j}, \Delta \phi_{1j}, \Delta r_{2j}, \Delta \theta_{2j}, \Delta \phi_{2j}, 
        \\ &\dots, \Delta r_{Nj}, \Delta \theta_{Nj}, \Delta \phi_{Nj})
    \end{aligned}
\end{equation}

\subsubsection{Spherical vector-based PSO algorithm}

The purpose of utilising spherical vectors in SPSO is to improve navigation safety by establishing a relationship between the speed, turning angle, and ascending angle of the UAV and the magnitude, elevation, and azimuth components of these vectors. As a result, the possibility of discovering excellent solutions is increased since the particles in SPSO investigate solutions inside the configuration space as opposed to the Cartesian space. Furthermore, the search space can be significantly reduced by explicitly incorporating limitations related to turning and ascending angles by adjusting the elevation and azimuth angles of the spherical vector. Fixing the magnitude can help further restrict the search space while increasing the search capacity in some circumstances, such as when the UAV stays at a fixed speed. The Pseudo-code for Static Optimal Path Planning using SPSO is in \cite{spso2021}.
% \newpage
\subsection{Environment Construction}
\begin{figure}[H]
  \centering

  \subfigure[3D view]{
    \includegraphics[width=0.4\textwidth]{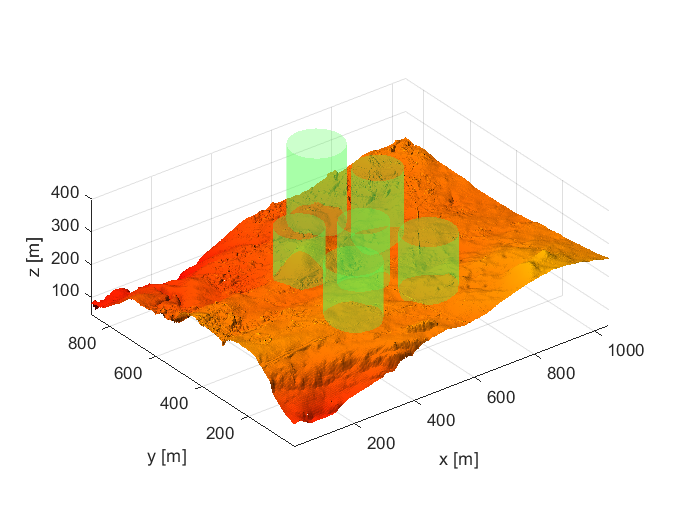}
    \label{1st_senario_3Dview}
  }
  \subfigure[Top view]{
    \includegraphics[width=0.4\textwidth]{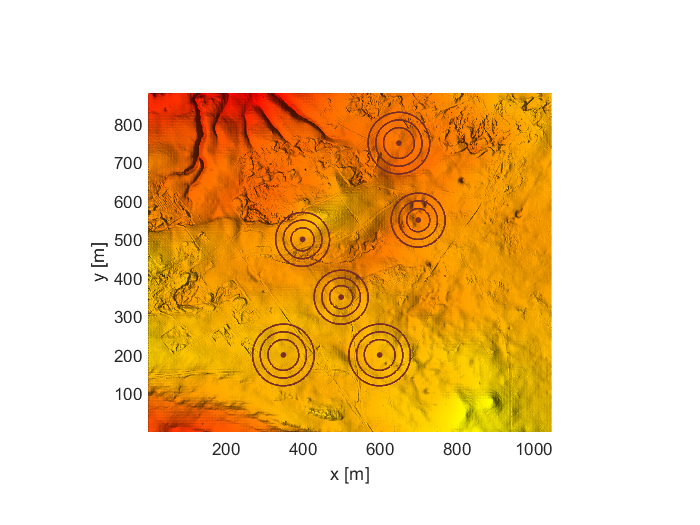}
    \label{1st_senario_topview}
  }
  
  \caption{Environment Isometric \& Top Views}
  \label{fig:EnvViews}
\end{figure}

The evaluation scenarios are constructed using actual digital elevation model (DEM) maps obtained from LiDAR sensors. Regions of Christmas Island in Australia, characterized by diverse terrain structures, are chosen \cite{spso2021}. Subsequently, these areas are enhanced to create different bench-marking scenarios, illustrated in this section. Within these scenarios, threats are depicted by green cylinders when seen in an isometric view, and as concentric circles when seen in a top view (see figure \ref{fig:EnvViews}).
\captionsetup{font={small}}

\subsection{Dynamic obstacles}
In addressing the dynamic path planning problem for the Unmanned Aerial Vehicle (UAV), way-points play a pivotal role as crucial checkpoints, discretizing the UAV's path through an environment teeming with randomly moving obstacles. In this section, the path planning challenge is dynamically tackled using an iterative solver operating online. This solver consistently updates the UAV's path in response to the dynamic nature of the environment, characterized by obstacles exhibiting random motion, necessitating real-time adjustments for effective UAV navigation.
The threat dynamics are simulated by random positional shifts, with each obstacle restricted to move in a circular area with a specified radius, and centered at its current position. All threats are assumed not to instantly jump on the UAV, making sure that the UAV does not fall inside its dead-zone. No threat is allowed to overlap with another threat as well.

The difficulty of the environment could be adjusted through changing both the threat size and the radii of the motion restricted areas for each threat.

\subsection{SPSO-DPP}
The SPSO dynamic path planner utilizes the SPSO-solver introduced by \cite{spso2021}. At each way-point, the solver reevaluates the environment, incorporating changes in obstacle positions. Notably, the SPSO-solver is employed at each way-point to ensure the path's coherence, optimizing it as the UAV progresses from one way-point to the next. This adaptive process unfolds dynamically, ensuring the UAV successfully navigates evolving scenarios, as visually depicted in figures \ref{fig:scenario1}, \ref{fig:scenario2}, \ref{fig:scenario3} and \ref{fig:scenario4}.
\captionsetup{font={small}}
\begin{figure}[H] \centering
    \includegraphics[width=0.35
\textwidth]{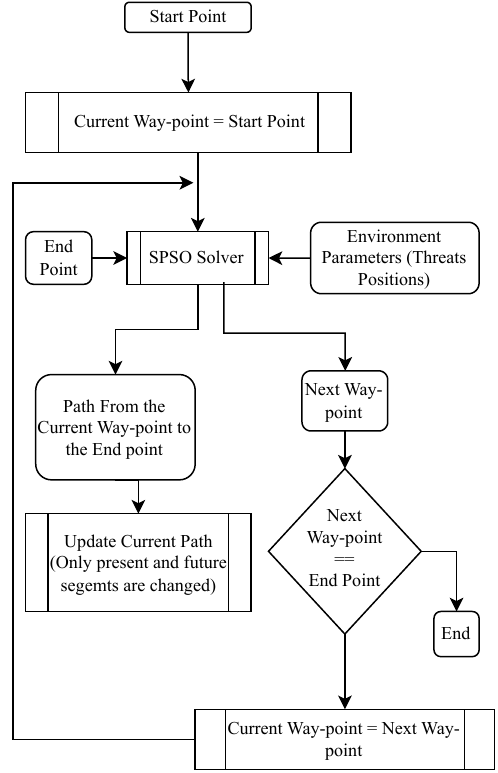}
    \caption{SPSO-DPP Flow Chart}
\label{fig:pdf} \end{figure}

\subsection{Path representation}
It is crucial to note that the UAV path is represented by a curve that is split into three sections: 1) a dashed-line section that resembles the temporary planned path from the next way-point to the destination; 2) a solid-line portion that represents the finalized path from the UAV's current way-point to the next way-point; and 3) a dashed-dotted-line portion that represents the path taken from the start point to the current point. The dash-dotted-line portion not only serves as a representation of the finalized path in space but also signifies the path that the UAV has previously traversed and is no longer subject to updates. Similarly, the solid-line portion denotes the current path taken by the UAV in the ongoing step or iteration and is not revisited for re-planning. In contrast, the dashed-line portion outlines the prospective path for all upcoming way-points, representing the segment that necessitates optimization and re-planning. This distinction in path portions ensures a focused approach, directing computational efforts towards refining the UAV's future path while preserving the integrity of its past and present routes. Within the overall path from the start to the destination, 10 way-points are positioned. Way-points 1 to 8 act as checkpoints for dynamic updates, where the solver re-calibrates the UAV's path. The last 2 way-points represent a final path segment, requiring no further re-planning as it constitutes a unique solution.

\section{Results}
\captionsetup{font={small}}
\begin{figure*}
    \centering
    \includegraphics[width=\textwidth]{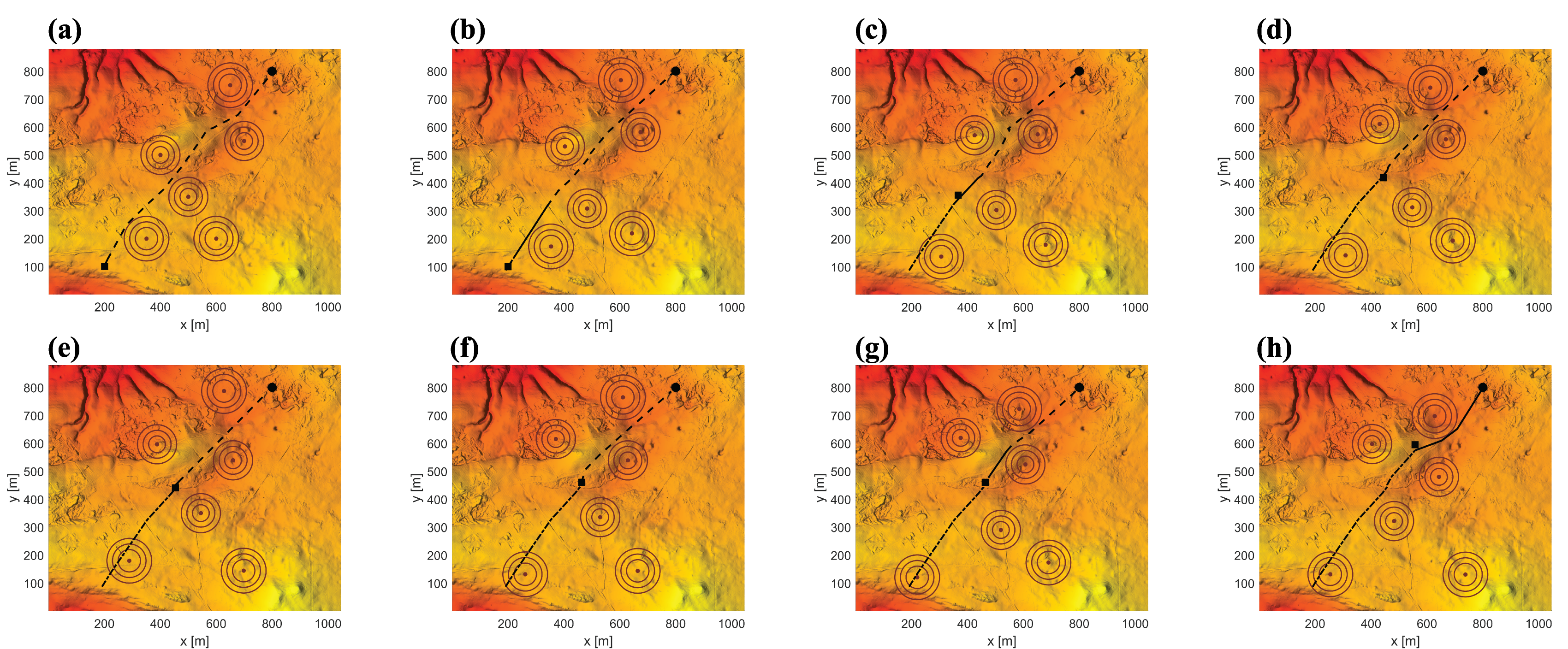}
    \caption{Base Case: Planned path at each way-point using the same weights as in \cite{spso2021}}
    \label{fig:scenario1}
\end{figure*}

\begin{figure*}
    \centering
    \includegraphics[width=\textwidth]{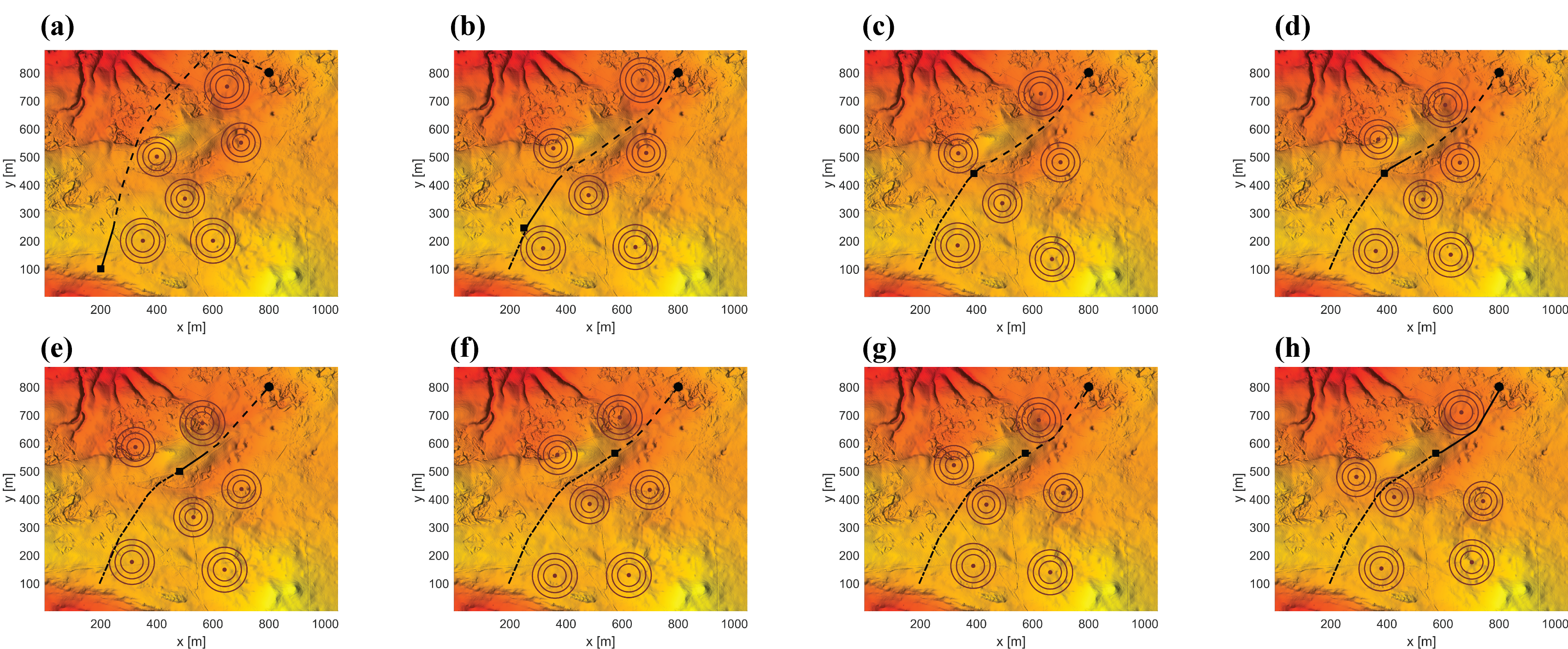}
    \caption{Safer Path Case: Planned path at each way-point with increasing the value of ${b_2}$ to 100 for ensuring a safer path for the UAV}
    \label{fig:scenario2}
\end{figure*}

\begin{figure*}
    \centering
    \includegraphics[width=\textwidth]{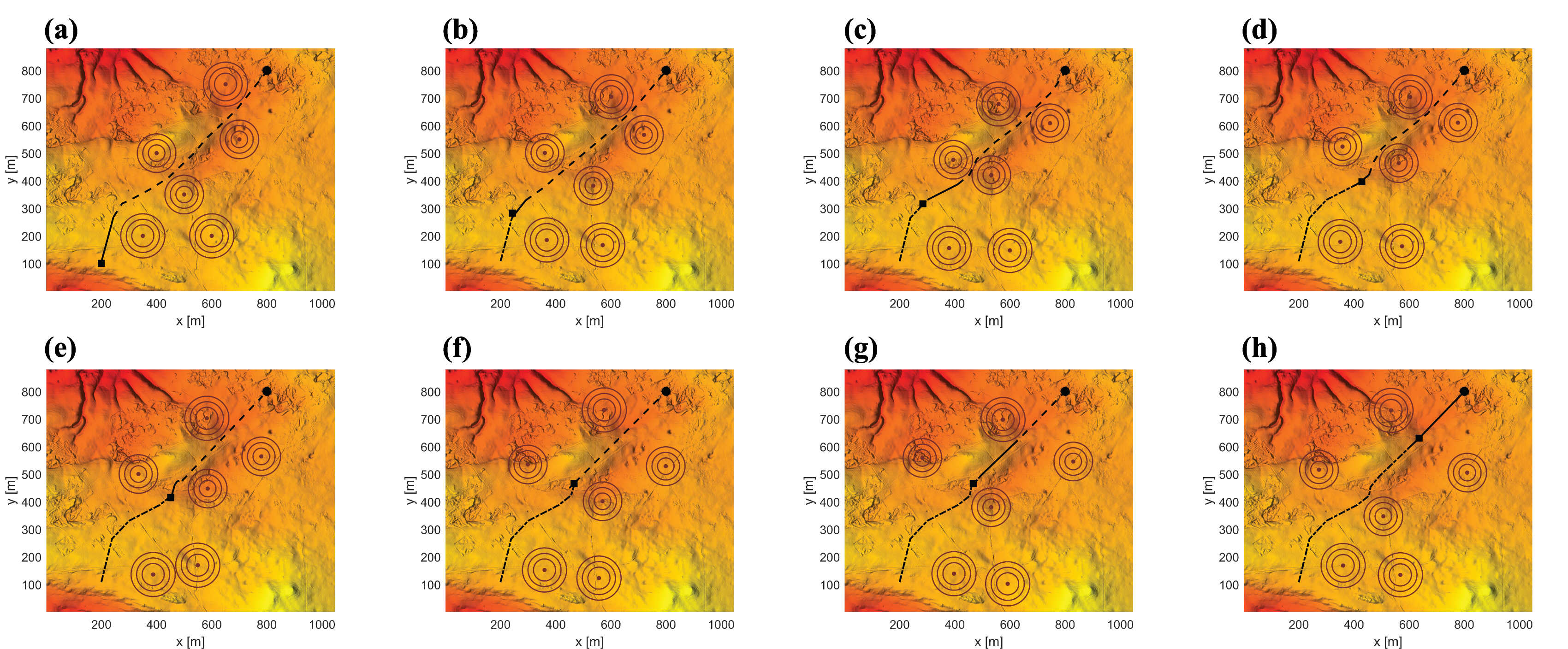}
    \caption{Shorter Path Case: Planned path at each way-point with decreasing ${b_2}$ to 1 for prioritizing the distance cost to achieve more dominance in the overall cost function}
    \label{fig:scenario3}
\end{figure*}

\begin{figure*}
    \centering
    \includegraphics[width=\textwidth]{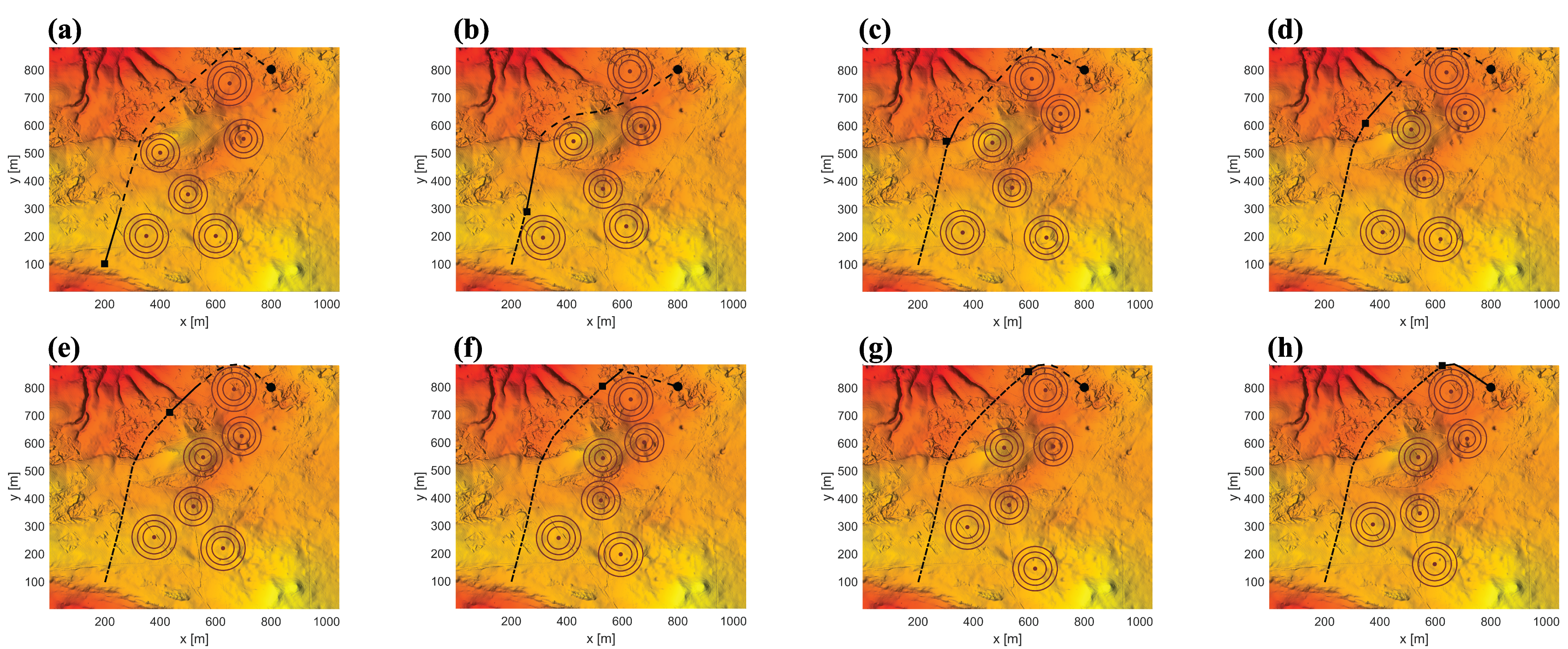}
    \caption{Safer Than Ever Case: Planned path at each way-point with increasing the value of ${b_2}$ to be 500 prioritizing safety over distance minimization for ensuring a higher level of safety for the UAV}
    \label{fig:scenario4}
\end{figure*}

\subsection{Case Scenarios}
Four case scenarios are presented to test the effect of cost weights, specifically the safety and distance weights.
\begin{itemize}
    \item Case 1 - "Base Case": The weights are taken to be similar to \cite{spso2021}. The parameters are set as follows: ${b_1}$ = 1, ${b_2}$ = 5, ${b_3}$ = 10, and ${b_4}$ = 1.
    \item Case 2 - "Safer Path Case": A modification is made specifically to ${b_2}$, increasing its value to 100. This adjustment is implemented as a means of ensuring a safer path for the UAV by elevating the threat cost coefficient. The impact of this change is evident in the planned paths showcased in Figures \ref{fig:scenario1} and \ref{fig:scenario2}, illustrating the discernible effect of altering the weights or importance assigned to the distance and threat objectives within the problem formulation.
    \item Case 3 - "Shorter Path Case": ${b_2}$ is decreased to 1; strategically prioritizing the distance cost to achieve more dominance in the overall cost function.
    \item Case 4 - "Safer Than Ever Case": ${b_2}$ value is increased to be 500 prioritizing safety over distance minimization. As a consequence, the algorithm naturally generates longer paths, reflecting a deliberate trade-off. This adjustment aims to ensure a higher level of safety for the UAV, even at the expense of producing relatively more extended and resource-intensive routes.
\end{itemize}
Each case is represented as 8 sub-figures resembling the past, present and future path portion at each way-point.

% \begin{twocolumn}

It is obvious from the comparison of the case 1 and case 2 that planned path tend have safer segments when giving the threat cost a higher weight. This can be explicitly seen in Figures \ref{fig:scenario1}-a and \ref{fig:scenario2}-a.

Despite of the closely position obstacles in case 3, the planned path have segments that go between these obstacles without much consideration of the high threat the UAV may face.  Figure \ref{fig:scenario3}-c is a resemblance of a relatively low threat cost effect with a high distance cost. In Figure \ref{fig:scenario3}-d, e, and f, another intriguing observation is that, at certain way-points, the solver opted for nearly non-progressive segments of very small lengths instead of choosing a longer route. This in a way shows the strength of the dynamic planner, where it is not just capable of determining the best route according to the momentary obstacles positions, but also, in future, may be further developed to account for the obstacles direction and speed of motion.

Aggressively increasing the threat cost resulted in a relatively longer planned route as seen in Figure \ref{fig:scenario4}. The solver did not go for a chance margin of collision, not even a small one. Undoubtedly, these observations strongly indicate a significant trade-off between safety and the distance of the path.

\subsection{Performance of SPSO-DPP against GA \& PSO}
The first three case scenarios in the previous section we re-run using SPSO-DPP in comparison to GA and PSO. Two tests were conducted: one where the random movement of the threats were held constant across all methods and scenarios, and another where the the randomness of the threats movements varied for each run. In the first test, the performance is presented as the direct comparative costs of each method, while in the second test, it is reported as the averaged costs from many runs for each scenario. Finally, an additional test is conducted to evaluate methods average performance across three different settings of threat dynamics.

The cost summery of all cases are tabulated and shown in Tables \ref{tab:cost_dynamic_comparison}, \ref{tab:threatvar} and Figure \ref{fig:avged}. The highlighted columns indicate that SPSO provide better results for the cost compared to PSO and GA. Cost 1 signifies the expense from the first way-point to the end of the path. Similarly, Cost 2 denotes the cost from the second way-point, where the path is recalculated, to the end of the path. This pattern continues for the subsequent costs, up to Cost 8, each reflecting the recalculated path from the corresponding way-point to the destination.

\subsubsection{First Test}
SPSO demonstrated superior performance compared to both PSO and GA. In comparison to PSO, SPSO exhibited cost reductions of up to 330\%, 340\%, and 632\% in cases 1, 2, and 3, respectively. When compared to GA, SPSO demonstrated cost reductions of up to 390\%, 380\%, and 675\% in cases 1, 2, and 3, respectively. Furthermore, it is noted that the application of the SPSO-solver leads to a consistently non-increasing trend in path cost as the UAV progresses through the way-points. This suggests that the SPSO-solver consistently identifies an optimal way-point for the current movement. In contrast, the GA exhibited fluctuations at various instances (e.g., Case 1 - costs 4 and 5, Case 2 - costs 1 and 2, and Case 3 - costs 3 and 4), indicating that the selected way-point by the algorithm may not necessarily be optimal.

\subsubsection{Second Test}
\begin{figure}[H]
    \centering
    \includegraphics[scale=0.6]{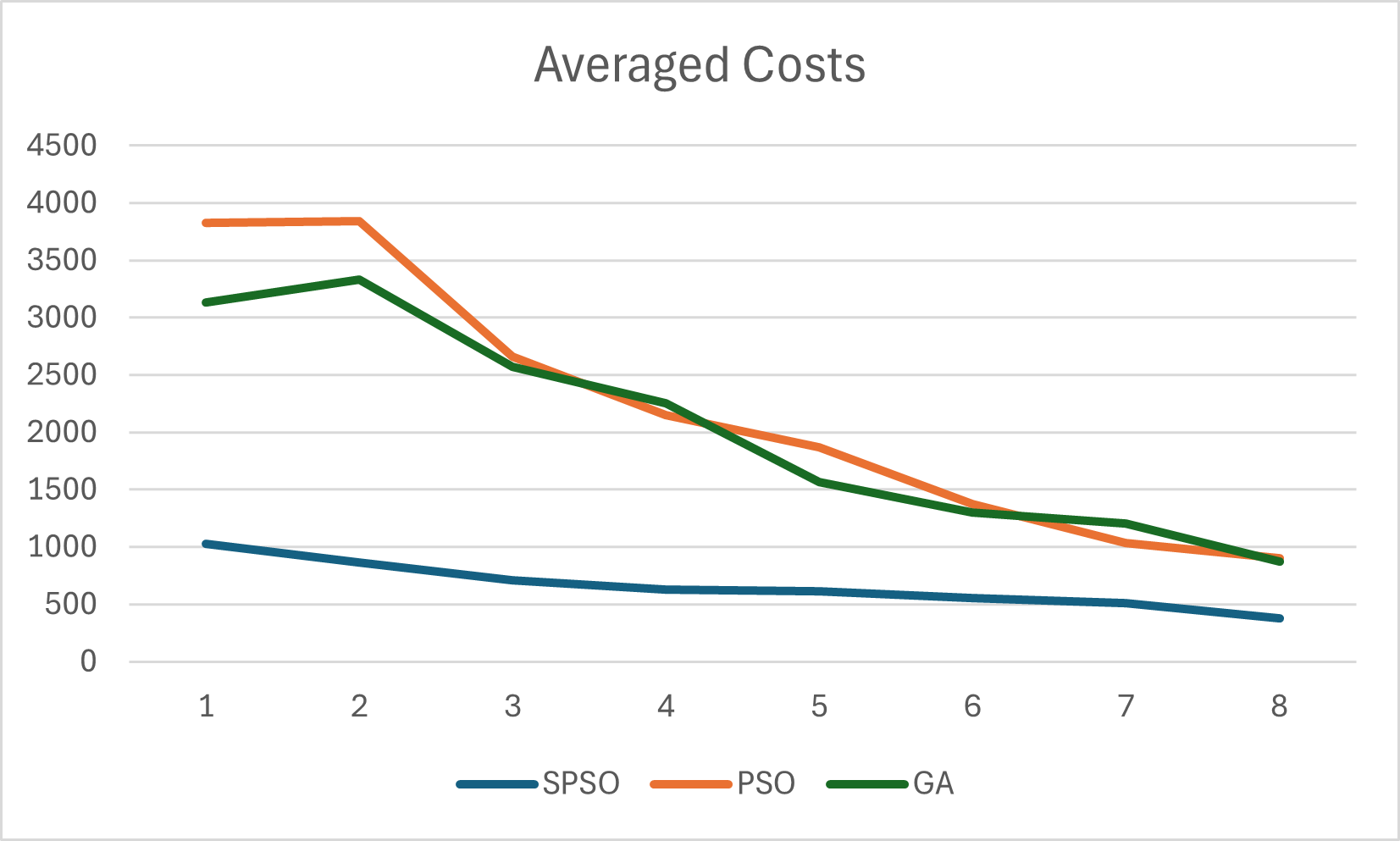}
    \caption{SPSO vs PSO vs GA Averaged Costs across all Scenarios}
    \label{fig:avged}
\end{figure}

Another evaluation metric for the SPSO-DPP algorithm is to get  averaged performance statistics for the algorithm compared to the state of arts GA and PSO by running it 10 times with random motion of the obstacles in each run and calculate the average cost over the total number of runs. Figure \ref{fig:avged} shows the averaged costs of each method across all scenarios and runs. Even though all methods find a solution path with an almost non-increasing cost, SPSO outperforms GA and PSO significantly in a dynamic environment on average.

%\onecolumn
\captionsetup{font={small}}
\begin{table}[H]
\centering
\resizebox{0.5\textwidth}{!}{
\begin{tabular}{@{}cccccccccc@{}}
\toprule
       & \multicolumn{3}{c}{Scenario 1} & \multicolumn{3}{c}{Scenario 2} & \multicolumn{3}{c}{Scenario 3} \\ \midrule
       & SPSO           & PSO   & GA    & SPSO           & PSO   & GA    & SPSO             & PSO  & GA   \\
Cost 1 & \textbf{1058}   & 3269  & 4424  & \textbf{946}  & 4173  & 3205  & \textbf{944}     & 4016 & 2837 \\
Cost 2 & \textbf{1013}   & 2866  & 3127  & \textbf{783}   & 2307  & 3504  & \textbf{637}     & 3493 & 3612 \\
Cost 3 & \textbf{790}   & 2626  & 2650  & \textbf{708}   & 2440  & 3405  & \textbf{477}     & 3247 & 2231 \\
Cost 4 & \textbf{655}   & 2512  & 1861  & \textbf{753}   & 2223  & 2704  & \textbf{368}     & 2697 & 2574 \\
Cost 5 & \textbf{492}   & 1695  & 2418  & \textbf{455}  & 1589  & 1793  & \textbf{237}     & 1673 & 1839 \\
Cost 6 & \textbf{410}   & 1273  & 1478  & \textbf{362}   & 1221  & 1135  & \textbf{226}   & 1375 & 1117 \\
Cost 7 & \textbf{346}   & 1038  & 933  & \textbf{281}   & 1029  & 800  & \textbf{226}   & 1098 & 1038  \\
Cost 8 & \textbf{184}   & 789   & 819   & \textbf{271}   & 843   & 787  & \textbf{226}  & 937  & 908  \\ \bottomrule
\end{tabular}}
\caption{SPSO vs PSO vs GA costs in Fixed Obstacles Motion}
\label{tab:cost_dynamic_comparison}
\end{table}

% \captionsetup{font={small}}
% \begin{table}[H]
% \centering
% \resizebox{0.5\textwidth}{!}{
% \begin{tabular}{@{}cccccccccc@{}}
% \toprule
%        & \multicolumn{3}{c}{Scenario 1} & \multicolumn{3}{c}{Scenario 2} & \multicolumn{3}{c}{Scenario 3} \\ \midrule
%        & SPSO           & PSO   & GA    & SPSO  & PSO   & GA    & SPSO             & PSO  & GA   \\
% Cost 1 & \textbf{1015.5} & 4080.7 & 3486.6 & \textbf{1019}  & 3379   & 2782   & \textbf{1046}  & 4010   & 3138 \\
% Cost 2 & \textbf{809.9}  & 3916.3 & 2874.9 & \textbf{888.3} & 3401.7 & 3454.8 & \textbf{906}   & 4217   & 3660 \\
% Cost 3 & \textbf{702.3}  & 2293   & 2342   & \textbf{696.5} & 2758.6 & 2878.4 & \textbf{730}   & 2917   & 2501 \\
% Cost 4 & \textbf{556}    & 1944   & 2372   & \textbf{675}   & 2291.5 & 2234   & \textbf{655}   & 2217   & 2155 \\
% Cost 5 & \textbf{525.4}  & 1712   & 1700   & \textbf{638.5} & 1671   & 1806   & \textbf{685}   & 2224   & 1184 \\
% Cost 6 & \textbf{417}    & 1337   & 1264  & \textbf{569}   & 1340.7 & 1447.3 & \textbf{682}   & 1449.8  & 1194 \\
% Cost 7 & \textbf{322}    & 1066   & 1234   & \textbf{546.8} & 1037   & 22613  & \textbf{658}   & 1003   & 1096 \\
% Cost 8 & \textbf{285}    & 854    & 871.7  & \textbf{284}   & 907    & 951    & \textbf{558}   & 938    & 804  \\ \bottomrule
% \end{tabular}}
%     \caption{SPSO vs PSO vs GA Averaged costs in Random Obstacles Motion}
%     \label{tab:randcomp}
% \end{table}

\subsubsection{Third Test}
Using the base case Scenario (Case 1), SPSO-DPP, PSO and GA were tested across various environment difficulties, simulated by varying the obstacles allowed motion radii. The radii are specifically set to 50, 100, and 200 in this test. Table \ref{tab:threatvar} shows the average costs of 10 runs for each radius value. Again, SPSO-DPP showed obvious domination over the other methods.

\begin{table}[H]
    \centering
\resizebox{0.5\textwidth}{!}{
\begin{tabular}{@{}cccccccccc@{}}
\toprule
       & \multicolumn{3}{c}{R = 50} & \multicolumn{3}{c}{R = 100} & \multicolumn{3}{c}{R = 200} \\ \midrule
       & SPSO           & PSO   & GA    & SPSO    & PSO   & GA    & SPSO             & PSO  & GA   \\
Cost 1 & \textbf{1032.0} & 4226.8 & 2678.4 & \textbf{967.1} & 3460.0 & 4266.3 & \textbf{952.0} & 3490.4 & 2796.2 \\
Cost 2 & \textbf{849.2} & 3764.4 & 3390.1 & \textbf{784.4} & 2780.0 & 3573.5 & \textbf{752.5} & 3070.0 & 4227.8 \\
Cost 3 & \textbf{676.6} & 2803.6 & 2515.0 & \textbf{665.3} & 2432.5 & 2468.6 & \textbf{640.6} & 2120.2 & 2946.8 \\
Cost 4 & \textbf{562.4} & 2041.1 & 2234.9 & \textbf{567.6} & 2516.3 & 2336.3 & \textbf{612.5} & 1790.5 & 1814.4 \\
Cost 5 & \textbf{507.1} & 1687.2 & 2007.5 & \textbf{491.9} & 1892.4 & 1695.6 & \textbf{652.9} & 1482.3 & 1758.3 \\
Cost 6 & \textbf{383.9} & 1232.9 & 1142.3 & \textbf{411.3} & 1225.2 & 1263.6 & \textbf{555.2} & 1382.0 & 1918.2 \\
Cost 7 & \textbf{343.0} & 1123.7 & 1097.7 & \textbf{348.9} & 1108.7 & 892.6 & \textbf{375.7} & 973.9 & 1484.6 \\
Cost 8 & \textbf{246.4} & 781.3 & 889.7 & \textbf{360.0} & 889.8 & 860.6 & \textbf{308.2} & 883.8 & 803.3 \\ \bottomrule
\end{tabular}}
    \caption{SPSO vs PSO vs GA Averaged costs Across Varying Threat allowed Motion Radii}
    \label{tab:threatvar}
\end{table}
% \begin{table}[H]
% \centering
% \begin{tabular}{@{}lllllllll@{}}
% \toprule
% Scenario 4 & Cost1 & Cost 2 & Cost 3 & Cost 4 & Cost 5 & Cost 6 & Cost 7 & Cost 8 \\ \midrule
%            & 1108  & 739    & 664    & 1329   & 3197   & 209    & 1316   & 143    \\ \bottomrule
% \end{tabular}
% \end{table}
\section{Limitations}
The presented work depends on the instantaneous relative threats positions, and does not take the velocity estimation of the dynamic threats into consideration. Thus, the SPSO-DPP does not account for the threats velocity vectors and their expected movement directions. Such estimation would enable further investigation of how the performance of SPSO-DPP is influenced by variations in environmental complexity.

\section{Conclusion}
Finally, the paper introduces the SPSO-DPP.
% The technique is utilized to address the complex path planning issues UAVs encounter in threat-filled, dynamic environments.
The approach uses an online solver to guide the UAV across areas with randomly moving obstacles by using way-points as checkpoints. At each way-point, the SPSO-solver is used to adaptively optimize the path, provide coherence, and make adjustments in real-time for effective UAV navigation.
Four scenarios have been presented, each of which modifies the parameters to reduce distance or prioritize safety. The results demonstrate that SPSO consistently produces superior cost values than both GA and PSO, demonstrating the algorithm's efficacy in generating optimal paths. The effect of prioritizing distance reduction above hazard avoidance is best shown in Case 3, where the decision leads to a more direct path that circumvents obstacle constraints. This realization emphasizes the necessity of carefully weighing objectives against constraints in situations including dynamic UAV path planning. 

In various conducted tests, the SPSO-DPP algorithm consistently performs better in path cost optimization than GA and PSO. Whether the approaches are directly compared, the costs are averaged over several runs, or various levels of obstacle motion are taken into account, SPSO-DPP performs better and always yields optimal solutions.
Furthermore, we intend to validate our algorithm's output with real-world simulations, to confirm that the algorithm operates as efficiently and robustly as possible under demanding and dynamic UAV conditions.
% \end{twocolumn}
% \newpage

\section{Acknowledgments}
{The authors would like to acknowledge the support of IRC for Sustainalbe Energy Systems through the funded project no. INRE2328. Dr. Abido acknowledges the support recieved from Saudi Data and AI Authority (SDAIA) and KFUPM under SDAIA-KFUPM Joint Research Center for Artificial Intelligence no. JRC-AIRFP-09.
}
% \section{Appendix}
% \newpage
\bibliographystyle{ieeetr}
\bibliography{References.bib}

\end{document}